# Fog Robotics for Efficient, Fluent and Robust Human-Robot Interaction

Siva Leela Krishna Chand Gudi, Suman Ojha, Benjamin Johnston, Jesse Clark, Mary-Anne Williams
Innovation and Enterprise Research Laboratory (The Magic Lab), Centre for Artificial Intelligence
University of Technology Sydney, Australia
Email: 12733580@student.uts.edu.au, {suman.ojha, benjamin.johnston, jesse.clark, mary-anne.williams}@uts.edu.au

*Abstract*—Active communication between robots and humans is essential for effective human-robot interaction. To accomplish this objective, Cloud Robotics (CR) was introduced to make robots enhance their capabilities. It enables robots to perform extensive computations in the cloud by sharing their outcomes. Outcomes include maps, images, processing power, data, activities, and other robot resources. But due to the colossal growth of data and traffic, CR suffers from serious latency issues. Therefore, it is unlikely to scale a large number of robots particularly in human-robot interaction scenarios, where responsiveness is paramount. Furthermore, other issues related to security such as privacy breaches and ransomware attacks can increase. To address these problems, in this paper, we have envisioned the next generation of social robotic architectures based on Fog Robotics (FR) that inherits the strengths of Fog Computing to augment the future social robotic systems. These new architectures can escalate the dexterity of robots by shoving the data closer to the robot. Additionally, they can ensure that human-robot interaction is more responsive by resolving the problems of CR. Moreover, experimental results are further discussed by considering a scenario of FR and latency as a primary factor comparing to CR models.

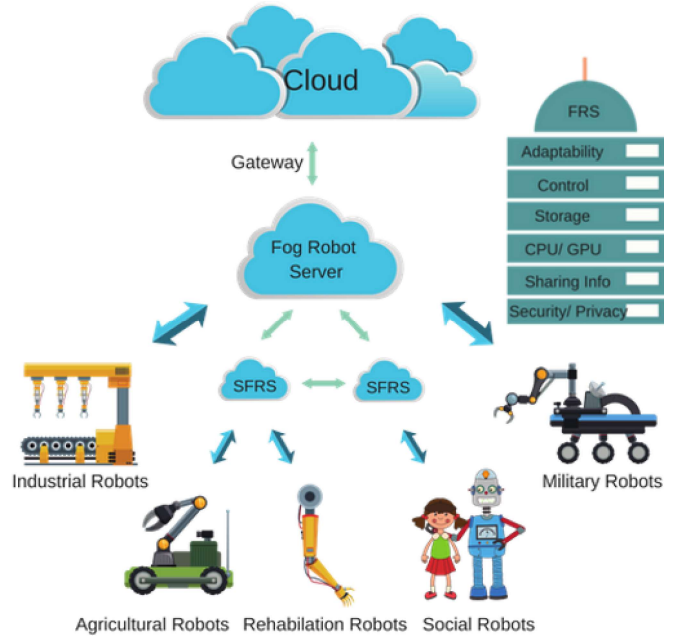

Fig. 1. Architecture of Fog Robotics.

## I. INTRODUCTION

Robots are playing a crucial role both in personal and social life [1] as well as in industries [2]. Notably, social robots such as Nao, Pepper, Paro, and Erica are emerging to support humans in a range of applications from helping autism patients to assisting older adults. To make these kind of robots smarter and more responsive to humans, Cloud Robotics(CR) was introduced to enable robots to share their outcomes such as updated libraries of maps, data, objects, processing power, images, and other robot resources [3]. CR working process includes requesting information, analyzing, interpreting, and responding back to the robot with a confirmation. As such, the operation of CR requires sending and receiving data over significant distances. Researchers expect that such data can grow to 2.3 zettabytes per year by 2020 [4]. They also claim that 50 billion devices are estimated to connect to the Internet by 2020 and more than 2 exabytes of data will be generated from billions of previously unconnected devices each day. This leads to an exponential growth in the demand for bandwidth for the fetching of data to and from cloud due to a large amount of generated traffic. As a result, cloud performance might degrade creating network congestion, latency, lower efficiency, and significant decrease in reliability. Additionally, as the robots surge in number, they too may start sending vast amounts of data to the cloud. Their architectures are not designed to tackle this growing volume of data and will encounter the above-mentioned issues.

*Latency* is one of the most critical problems of CR for human-robot interaction. When working with robots, humans expect them to execute actions in near real-time. A lag in command reception by a robot can cause the robot to do undesirable and awkward actions such as replying late or performing an unintended task which is not needed at the particular moment of time. Moreover, latency delays in robot actions can cause significant safety issues for humans in close proximity of the robot. In hard real-world applications, robots can also be damaged or may collide due to a lag in the detection of obstacles. Even a lag of milliseconds may slow robot's responses to the user, and therefore negatively impact the user experience.

This research was supported by the ARC, Innovation and Enterprise Research Laboratory (The Magic Lab) and Centre for Artificial Intelligence of the University of Technology Sydney, Australia.

CR architectures are not efficient enough for human-robot interaction, particularly in complex and dynamic environments. Few companies are already providing additional GPU for robots for performing artificial intelligence/machine learning (AI/ML) algorithms to handle the issue of latency. For example, Nao robots are using an extra backpack that contains additional GPU for efficient processing of complex algorithms [5]. However, these techniques are not sustainable when it comes to managing the future traffic because of large data. Therefore, we propose a novel social robotic architecture based on Fog Robotics (FR) using the well-established concept of Fog Computing [6], which not only solves the issue of latency inherent with CR but also offers better security, privacy, and robot collaboration.

Our main contribution of the paper lies in showing the proposed models of next-generation FR architecture and validating them on how it affects the response rate of robots. We specifically focus on the aspect of latency to compare FR and CR cases for conveying the importance of FR. In this paper, we first provide a brief introduction about Fog Robotics in Section II. In Section III, we discuss the comparison of Fog and Cloud Robotics. Section IV will present three different architectures of FR along with a scenario of *Delivery Social Robots*. Next, we present the results related to performance evaluation validated by simulation experiments in Section V and concludes the contributions of the paper in Section VI.

## II. FOG ROBOTICS

Fog Robotics can be defined as an architecture which consists of storage, networking functions, control with decentralized computing closer to robots. S.L.K.C. Gudi et al., proposed the idea of FR and scenarios with possible applications [7]. Fig. 1 shows the basic illustration of FR. It consists of Fog Robot Server (FRS) and a Sub Fog Robot Server (SFRS). FR extends the cloud along with computations near to the user with the help of FRS. FRS/SFRS are adaptable, consists of processing power for computation, network capability, and secured by sharing the outcomes of robots to other robots for efficient performance with better response rate. Storage, CPU/GPU can be permanent or temporary while it changes upon the necessity. The working process of FR and functions of FRS are as shown below:

### A. Working Methodology of Fog Robotics

- Robot makes a request for information to the FR system
- Request handovers to SFRS
- Upon analyzing, if SFRS is capable of responding the request, it can pass on to the robot
- Else SFRS seeks FRS to process the request
- If FRS too is unable to handle the request, it seeks the help of cloud
- Cloud solves the request and passes the information to robot

### B. Functions of Fog Robot Server

- Receives request from robot and aggregates the data
- Regularly send data summaries to cloud
- Upon demand of data, cache temporarily for future use
- Suggest for additional deployment of SFRS
- Maintain enhanced security/privacy
- Can be remotely operable in case of failure

Moreover, if the traffic is inflated at a particular area then SFRS can be launched. The functions of SFRS is similar to FRS. But it covers a smaller area by maintaining a data log provided by FRS. To maintain a better network between the systems, analytics can be used for optimization of FR systems such as traffic, bandwidth usage, failure prediction, and recommendation depending on previously utilized data. Based on the analytics received, FRS can be operated at specific locations with different topologies such as bus, star, mesh, and tree. They can be deployed quickly anywhere when they are in need due to its portability. Examples include shopping malls, parks, universities, hotels, restaurants, and a power pole on roads. In the upcoming section, differences between Fog and Cloud Robotics are further discussed for effective understanding.

## III. COMPARISON BETWEEN FOG AND CLOUD ROBOTICS

Fog Robotics (FR) is inspired by the concept of Fog Computing (FC). Having concerns with the hitches of Cloud Computing, CISCO coined the term FC which has recently emerged as an alternative to solve some of the prominent hurdles related to IoT, 5th generation wireless systems (5G), Healthcare, and Vehicular computing. However, FR shares some of the characteristics of FC. For example, they include the placement of fog nodes/servers and low latency communication but differs mainly in GPU/CPU power, storage, mobility and unacceptable real-time low latency interaction as the requirements are completely divergent to robotics field. On the other hand, Cloud Robotics (CR) was introduced by J. Kuffner invoking cloud technologies such as Cloud Computing and Cloud Storage [8]. Regardless, both are intended to share the outcomes of the robot but with some differences.

FR can be independent of cloud and approaches cloud only if in case it cannot acquire information from FRS for the robot. Compared to the storage of data, distributed FR maintains smaller or transient adaptable memory relying on the requirement/traffic of different situations while CR has a higher permanent memory. To access the data in FR, the robot can use FRS or cloud making the decision locally thereby consuming less on-board CPU/GPU power. It maintains a high response rate within milliseconds as it mostly consists of only one hop with local coverage. However, in case of cloud, there is only one way for the centralized cloud to access data and this makes the response time lower with high latency as the coverage is for a whole state or a country. Further differences between the Fog Robotics and Cloud Robotics are briefly summarized in Table 1.

Despite their differences, both have a similarity of sharing the same applications, environment models, outcomes, and transfer of heavy computation to the server. But the main problem of CR is that it can make a robot to stop working suddenly when it can not access data from the internet due to a burden on fronthaul/backhaul, while in FR, it is quite unlikely to happen. Moreover, FR can have specific high standard secured encryptions/protocols to keep it away from ransomware attacks with low power consumption. Besides, it can also improve the battery life of the robot. On the other hand, CR is prone to a high number of attacks due to its sharing of common protocol along with higher power consumption rate. On the assumption of a hack, a specific affected FRS will be made to shut down while others continue to work as usual and in CR, there is a need to halt the whole system. Furthermore, in contrast with CR, robots can become cheaper using FR because there is no need of using higher computation or powerful hardware in the robot such as the processing of AI/ML algorithms. The significance of FR with respect to CR can be further analyzed using a scenario of Delivery Social Robots after the discussion of proposed architectures.

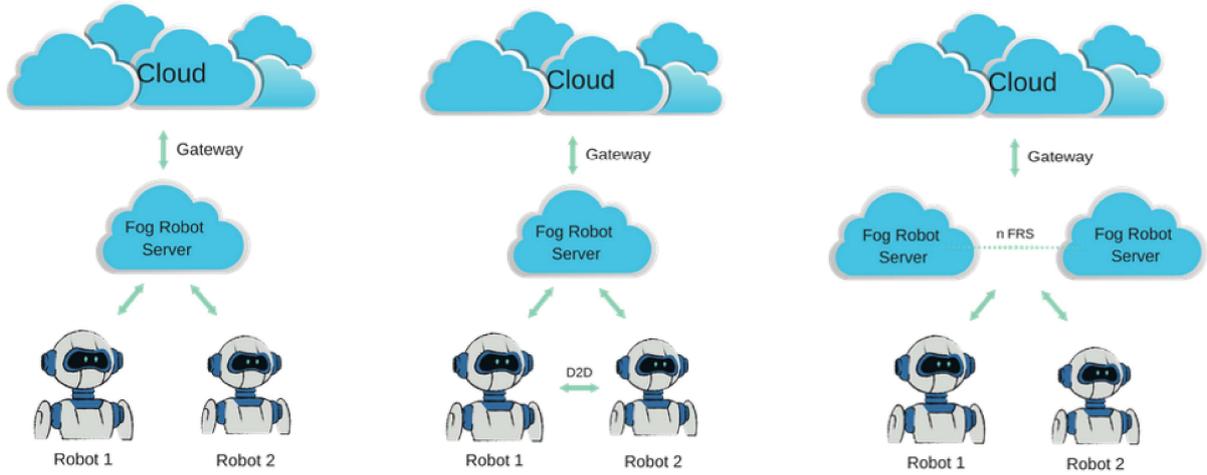

Fig. 2. Case A) Basic FR Architecture, Case B) FR Architecture with D2D Communication, Case C) FR Architecture with Multiple Fog Robot Servers

TABLE I
COMPARISON BETWEEN FOG AND CLOUD ROBOTICS

| Parameters | Fog Robotics | Cloud Robotics |
| --- | --- | --- |
| Storage | Low | High |
| Storage Type | Transient | Months/Years |
| Location | Distributed | Centralized |
| Decision Making | Local | Remote |
| CPU/GPU | Low | High |
| Response Time | Milliseconds | Seconds |
| Topology | Mostly one hop | Multiple hops |
| Coverage | Local | State/country |
| Latency/Jitter | Low | High |
| Burden on Fronthaul | Low | High |
| Security protocols | Specific | General |
| Power Consumption | Low | High |
| Applications | Robots, Humanoids | Robots, Humanoids |

## IV. PROPOSED ARCHITECTURE

In this section, we propose next generation models of three different architectures in Fig. 2 based on Fog Robotics. In all of the three architectures, Fog Robot Server support the cloud and contacts cloud only if in fact there is no information available at FRS level. The proposed architectures are as shown below.

### A. Basic FR Architecture

In this architecture, the FR system consists of a single Fog Robot Server with multiple robots. Robots share information and communicate each other with the help of FRS. If the number of robots increases, then the configuration of FRS can be expanded to manage the additional incoming data. It also depends upon the traffic and the type of data that is being used. Anyway, the performance degrades when number of robots are utilizing a single FRS because there is a limit for monitoring the robots. When this situation occurs, it shifts to either second or third architecture (to be discussed in the following section). This kind of architecture can be applied for a few number of robots deployed in home, restaurants, and banks.

### B. FR Architecture with D2D Communication

In this architecture, in addition to a single Fog Robot Server, a device to device communication (D2D) is added. This makes the robot communicate among themselves which belong to a similar area. As this architecture does not involve the Fog Robot Server, the performance rate i.e., the transfer of data and learning by the robot is faster than the first architecture which in turn minimizes the latency. This can be utilized when the robots are near to each other, and this method becomes void when they are far. In distant case scenario, it immediately starts using FRS to continue its activities. This model can be implemented for robots used in homes and hotels for enabling collaboration/communication between them.

## C. FR Architecture with Multiple Fog Robot Servers

In this architecture, the number of Fog Robot Servers are extended in a particular locality. This can boost the performance of robot in a better way than the previous systems. A multiple number of robots can be used in this configuration. It can also manage a tremendous amount of traffic. FRS communicate with the nearby FRS for sharing of data. After confirmation of unavailability of required data from the network of FRS, it triggers to the cloud for further process. This kind of architecture can be applied in future because there could be a lot of robots working and collaborating each other with the backing of multiple FRS. This architecture can be suitable for the robots in situations such as airports and parks.

For further demonstrating the importance of FR, we considered Delivery Social Robots scenario in the next section.

### Scenario of Delivery Social Robots

Delivery Social Robots (DSR) move around from one region to another for delivering goods as well as to have a chat with a human. On its way to delivery, their tasks involves to recognize obstacles, analyze images/maps, voice interaction, updating its current location to the server, collaborating with other DSR or to understand the present situation on its goal to the destination. To accomplish their objectives, they mostly rely on camera and a database related to voice interaction, maps, and image classification. On-board memory is mostly not enough to calculate such kind of data. Moreover, AI/ML methodologies are generally used for accurate results. This requires higher computation power for analysis. So currently, it requests most of the above said information from CR for processing. For example, images are taken using the camera and sent to the cloud for further classification. Cloud verifies the received image and updates back to the DSR. Finally, DSR takes an appropriate action to proceed for the next step.

In most of the cases, the objects that have to be defined by DSR are movable at a faster rate. Due to this reason, the response rate must be higher with less latency to succeed its task. But latency is likely to increase because they generate a massive amount of data collected from FHD, UHD, QHD cameras, global positioning system, and various other sensors for processing in the cloud increasing the burden on fronthaul and backhaul. In addition, as the number of DSR increases, CR is not capable of handling this situation.

To tackle these issues, an FR system can be used. Any of the three architectures discussed earlier can be applied based on the requirement. FRS is loaded with the capability of storage, local information such as maps, tower/high building areas, trees to avoid a collision, details of local shops/restaurants, alternative paths, checkpoints, and danger zones. Additional GPU can be provided for the FRS to compute high-level AI/ML algorithms and other cognitive services if required. As cloud servers are located somewhere far away while FR is close to DSR, it can provide faster and accurate results by optimizing network usage even for high quality videos. Regardless of connection to the internet, DSR can carry on its task as usual within its network. Most of the functions are done offline, but it can connect to cloud if it cannot retrieve information from FRS. If in case, the traffic of DSR is increasing at some particular places then an additional SFRS can be added.

The advantages of FR over CR for DSR are that it provides low latency, faster and accurate results by making the robot always connected and executing its assigned tasks on time. It can also enhance the battery life of DSR. Therefore, overall QoS can be improved by providing scalability and flexibility. This can also make the robot cheaper by being small in size, using smaller hardware devices as it does not need higher computation power.

## V. RESULTS

In this section, we discuss the simulation platform and the measurement criteria considered. Subsequently, we show the simulation results by examining latency for different architectures. For evaluating the performance of FR, we chose iFogsim toolkit [9] which is derived from Cloudsim toolkit as a platform. FR scenario is simulated by creating the FRS and robots using the three architectures explained earlier. Generally, the data used in between Cloud, FRS, and DSR is mostly maps, video, speech processing, image analysis and AI/ML tasks. So, we consider this as raw data. This data must be sent through different channels of the system. The value of latency varies for all kinds of applications depending upon the task provided. So, we considered an assumption that there is a delay of 130ms for connection between Cloud and Gateway, from Gateway to FRS as 4ms, and FRS to DSR as 3ms while for internal execution process of the received data by robot 3ms of latency is maintained.

Coming to architectures A and B, one Fog Robot Server is considered with the number of DSR increasing from 1 to 5. Latency is measured when a multiple number of robots tried to access data using FRS, Cloud and D2D communication. For ease of understanding the comparison, both architectures are plotted in a single graph as the cloud generates the same result irrespective of the architectures due to the same configuration and latency parameters. Results say that for D2D, the latency increased from 3.8ms to 6.75ms with a transfer of information latency between the robots considered to be 2ms. For FR, it raised slowly from 8.82ms to 10.96ms and suddenly to 19.75ms concerning one to five robots. This sudden increase says that it reached the capacity of FRS. To further monitor this kind of latency, an extra SFRS can be introduced. Finally, Cloud showed a minimal fractional increase of latency from 276.58ms to 276.97ms for robots 1 to 5. This has not changed much because the cloud has the capacity to monitor more than five robots. But the latency is higher when we compare to the first two architectures of FR as the cloud is far away from the robot. Results of both cases are as plotted in Fig 3.

On the other hand of architecture C, there is a need to connect to multiple FRS with a number of robots so that robots can communicate each other within multiple FRS. So, we considered various FRS ranging from two to twenty with each handling four DSR. For better understanding and calculation of results, we chose 2, 5, 10, 15, 20 FRS. Based on the results achieved, we can say that FR system maintained a latency of 10.967ms even though the number of robots and FRS are increasing. This happened because FRS are near to the robot and are capable of processing all the four DSR. CR scenario maintained latency from 277.27ms to 278.47ms until five number of FRS. It started rising at an alarming rate as the load on cloud began to rise with 2126.52ms, 3152.94ms, 3666.07ms with respect to 10, 15, 20 number of FRS. These results suggest that a burden on the cloud is increasing when the number of robots/FRS rises in number which in turn increases the latency. The results of architecture C are as plotted in Fig. 4.

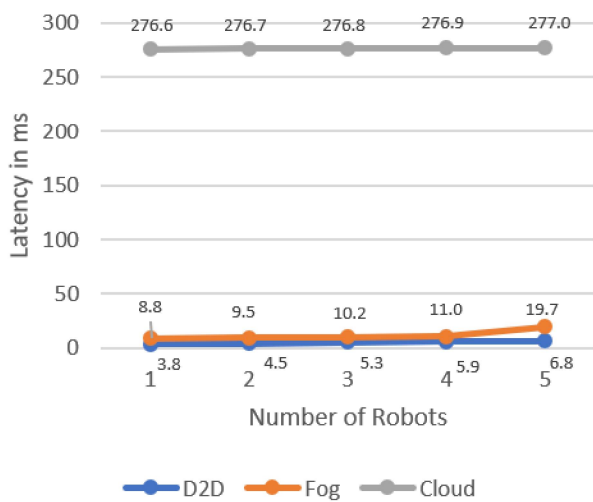

Fig. 3. Results of Architecture A and Architecture B Scenarios.

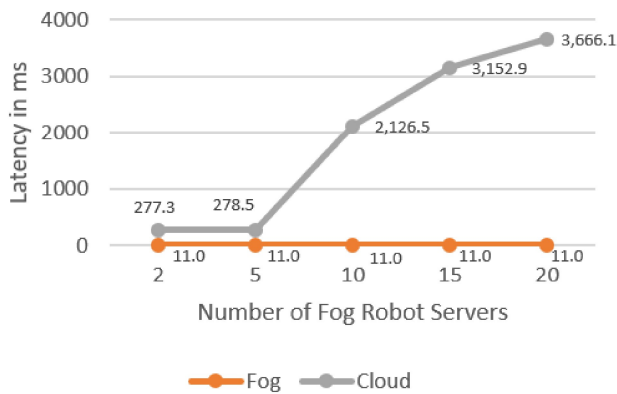

Fig. 4. Results of Architecture C Scenario.

If in case, all different latencies of a robot are considered together then it can rise at an alarming rate. Generally, this kind of latency can be easily observed on social robots when they take a long time to recognize human speech or while a robot is doing other tasks such as moving arms, head, legs, and other types of recognition. In summary, based on our results obtained we can conclude that using an FR scenario can make a lot of influence on latency. Higher latency by a DSR can make user feel annoying because it is not able to perform the task on time in addition to doing unwanted actions. So, FR can make a better impact for robust human-robot interaction.

## VI. CONCLUSIONS

In this work, we first gave a brief introduction of Fog Robotics for assisting the stand-alone robots. Then for better understanding, a comparison between Fog and Cloud Robotics is explored. Later, three different architectures of FR along with a scenario is discussed. Upon choosing latency as an aspect to validate FR, we successfully showed the results which proved that FR is far better than CR. It can become a companion to CR or work independently for intensifying efficient, fluent and robust human-robot interaction. Network bandwidth can be saved by processing the data locally using FRS. FR tackle data by reducing the burden on the cloud and process real-time communications by being decentralized and improving QoS. It reduces latency/jitter, eliminates bottlenecks caused by centralized systems, more secure by protecting sensitive data, increases the collaboration between robots, and better accuracy. Cheaper robots can be made because there is no need for a robot to have higher power processing capability with expensive hardware. FR architecture and scenarios can further be extended to different robotic applications such as medical, health care, industry, rehabilitation and player robots for better performance where the latency is considered as high priority. Ultimately, robots soon can be able to assist humans with their impressive performance. It can make the customers meet their expectations. For future research, we aim to extend our work by validating additional real-time scenarios and considering more functions on robots to further analyze the significance of Fog Robotics.